\documentclass{article} 
\usepackage{iclr2015,times}
\usepackage{hyperref}
\usepackage{url}

\usepackage{graphicx}
\usepackage{amsfonts}
\usepackage{amsmath}
\usepackage{mathtools}
\usepackage{bm}
\usepackage{bbm}

\DeclareMathOperator*{\argmin}{argmin}
\DeclareMathOperator*{\argmax}{argmax}

\title{Compact Compositional Models}

\author{
Marc Goessling\\
Department of Statistics\\
University of Chicago\\
Chicago, IL 60637, USA\\
\texttt{goessling@galton.uchicago.edu}
\And
Yali Amit\\
Departments of Statistics and Computer Science\\
University of Chicago\\
Chicago, IL 60637, USA\\
\texttt{amit@galton.uchicago.edu}
}

\iclrfinalcopy

\begin{document}

\maketitle

\begin{abstract}
Learning compact and interpretable representations is a very natural task, which has not been solved satisfactorily even for simple binary datasets. In this paper, we review various ways of composing experts for binary data and argue that competitive forms of interaction are best suited to learn low-dimensional representations. We propose a new composition rule that discourages experts from focusing on similar structures and that penalizes opposing votes strongly so that abstaining from voting becomes more attractive. We also introduce a novel sequential initialization procedure, which is based on a process of oversimplification and correction. Experiments show that with our approach very intuitive models can be learned.
\end{abstract}

\section{Introduction}
In recent years, multi-layer network architectures have drastically improved the discriminative performance in several classification tasks. These deep networks can be constructed based on desired invariances and stability properties \citep[e.g.,][]{bruna2013invariant} or they can be learned from usually large datasets as a cascade of nonlinear functions \citep[e.g.,][]{lee2009convolutional}. While being effective for classification the used transformations are typically high-dimensional and individual features are not always semantically meaningful. Moreover, the described structures are sometimes more global than desired and often there are many features that describe similar structures. As pointed out by \citet{bengio2013representation} learning to disentangle the factors of variation remains a key challenge in deep networks. Even for very simple classes, the basic task of learning a compact and interpretable representation has not yet been solved in a satisfactory manner. For example, the most natural representation of the letter T is in terms of a vertical and a horizontal bar. Consequently, this class can efficiently be represented by six coordinates, corresponding to the row/column location and the orientation of the bars.

In this work, we learn robust representations by seeking a parsimonious set of experts corresponding to the largest stable structures in the data. Apart from being intuitively appealing, low-dimensional explicit representations are, for example, useful for obtaining coarse scene descriptions in computer vision. We emphasize that our focus is not on achieving state-of-the-art performance in terms of classification rates or likelihoods but rather on learning simple models from few examples. In Section \ref{sec:composition_rules} we review various ways of composing experts for binary data and discuss their impact on the resulting representation. A new composition rule is then introduced that is particularly well suited for learning compact representations. The rule causes extremal competition among the experts over the data dimensions. In particular, the rule ensures that identical experts cannot be used to improve likelihoods and that opposing expert opinions always lead to a reduced likelihood.
In Section \ref{sec:sequential_inference} we present an appropriate sequential inference procedure for the type of compositional models that we consider. In Section \ref{sec:learning} we describe a batch as well as online version of an EM-style learning procedure for the new composition rule. Moreover, we propose a sequential initialization method, which can be described as a process of oversimplification and correction. Results for a synthetic dataset and handwritten letters are presented in Section \ref{sec:experiments}.

\section{Composition rules}
\label{sec:composition_rules}

We model binary data $\bm{x} \in \{0,1\}^D$ through a product Bernoulli distribution $\mathbb{P}(\bm{x} \,|\, \bm{\mu})$, i.e., the variables $\bm{x}(d)$ are assumed to be conditionally independent given $\bm{\mu}$. The global template $\bm{\mu}$ is a composition of experts $\bm{\mu_1},\ldots,\bm{\mu_K}$, where each expert is a Bernoulli template $\bm{\mu_k} \in [0,1]^D$. The way in which the experts are combined in order to create the composed template is specified through a \emph{composition rule} (also known as patchwork operation, mixing function or voting scheme). Such rules are the generative counterpart of activation functions in feed-forward neural networks. Formally, a composition rule is a function $\gamma\colon [0,1]^K \to [0,1]$ with a varying number $K$ of arguments. The composed template $\bm{\mu}$ is obtained by applying the composition rule to the expert ``opinions'' for each dimension
\[
\bm{\mu}(d) = \gamma(\bm{\mu_1}(d), \ldots, \bm{\mu_K}(d)).
\]
Of special interest is the ability of an expert to abstain from voting. By that we mean there exists a value $q \in [0,1]$ such that
\[
\gamma(q,p_1,\ldots,p_K) = \gamma(p_1,\ldots,p_K)
\]
for all $p_1,\ldots,p_K \in [0,1]$ and all $K$.

In the following we consider two classes of models. The first class is a natural choice for binary images and is referred to as write-black models \citep{saund1995multiple}. In these models the default variable state is ``off'' (white pixel). Underlying factors are now able to turn variables ``on'' (black pixel). If multiple causes for a variable are present the state will still be ``on''. This type of model is appropriate, for example, for figure-ground segmentations (white corresponding to background and black corresponding to foreground) where experts describe object parts. Composition rules for write-black models encode ``no vote'' through $\bm{\mu_k}(d)=0$ and we refer to them as asymmetric rules. Note that the value 0 is used for regions far away from the expert support. A template probability $\bm{\mu_k}(d)=1/2$ on the other hand is used for variables that are close to the boundary of the support and can hence be interpreted as ``not sure''. In write-black models, samples from individual experts will look like actual object parts.

The second class we consider are write-white-and-black models \citep{saund1995multiple}. In such models experts are able to cast votes in favor of ``on'' as well as ``off''. Composition rules for write-white-and-black models encode abstention through $\bm{\mu_k}(d)=1/2$ and we refer to them as symmetric rules. Note that ``not sure'' is also encoded through $\bm{\mu_k}(d)=1/2$, so this value can have either of the two meanings. Samples from single experts for image data will not look like actual object parts because about half of the pixels in the background region will be turned on.

\subsection{Asymmetric rules}

\subsubsection{Noisy or}
A straightforward composition rule for write-black models is the disjunctive composition
\[
\gamma(p_1,\ldots,p_K) = 1 - \prod_{k=1}^K (1 - p_k).
\]
The composed probability is simply the probability of observing at least one success when drawing independently from Bernoulli distributions with probabilities $p_1,\ldots,p_K$. This rule was used by \citet{saund1995multiple}.

\subsubsection{Sum of odds}
It was argued by \citet{dayan1995competition} that the noisy-or rule offers little incentive for experts to focus on different structures in the data. A more competitive composition rule was therefore proposed, which is of the form
\[
\gamma(p_1,\ldots,p_K) = 1 - \left(1 + \sum_{k=1}^K \frac{p_k}{1-p_k}\right)^{-1}.
\]
This rule can be motivated as the probability of observing a success when drawing independently from Bernoulli distributions with probabilities $p_k$ conditioned on observing at most one success. It is easy to see that the composed odds are just the sum of the individual odds. While in a global mixture model exactly one component is responsible for generating the whole data vector, here exactly one expert is responsible for each dimension that is turned ``on''. So, in contrast to the noisy-or rule the responsibility for individual variables is not shared.

\subsubsection{Maximum}
The most extreme form of competition is achieved through the max rule
\[
\gamma(p_1,\ldots,p_K) = \smashoperator[r]{\max_{k=1,\ldots,K}} p_k.
\]
Such a rule was used by \citet{lucke2008maximal}. Since only the strongest template matters, experts have no incentive to represent structures that are already present, unless their opinion is the most extreme one. Consequently, experts tend to focus on different aspects of the data. In contrast to the noisy-or and the sum-of-odds rule, likelihoods cannot be improved by using the same expert multiple times. Other than for the sum-of-odds rule, with this composition rule for each variable $\bm{x}(d)$ it is known which expert is responsible. This fact makes it possible to use an analytic formula in the M-step of the EM learning procedures for such models.

\begin{figure}[t]
\centering
\includegraphics[width=.495\textwidth]{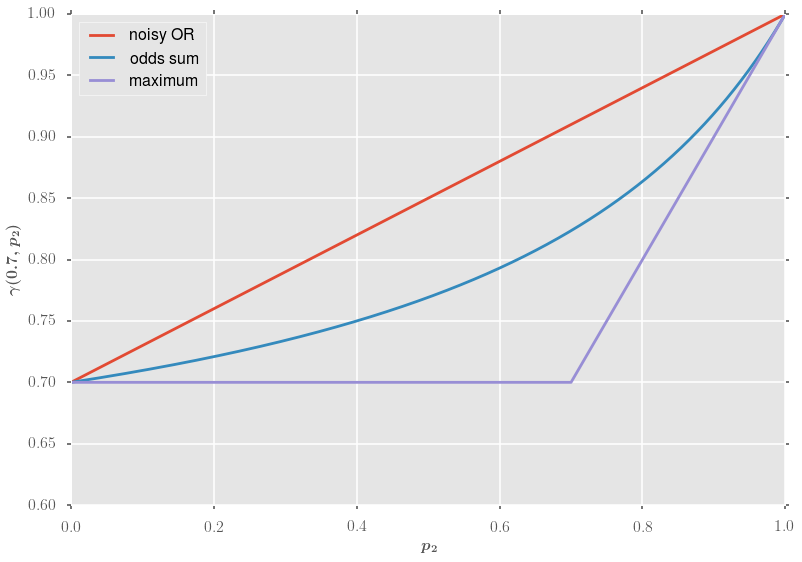}
\includegraphics[width=.495\textwidth]{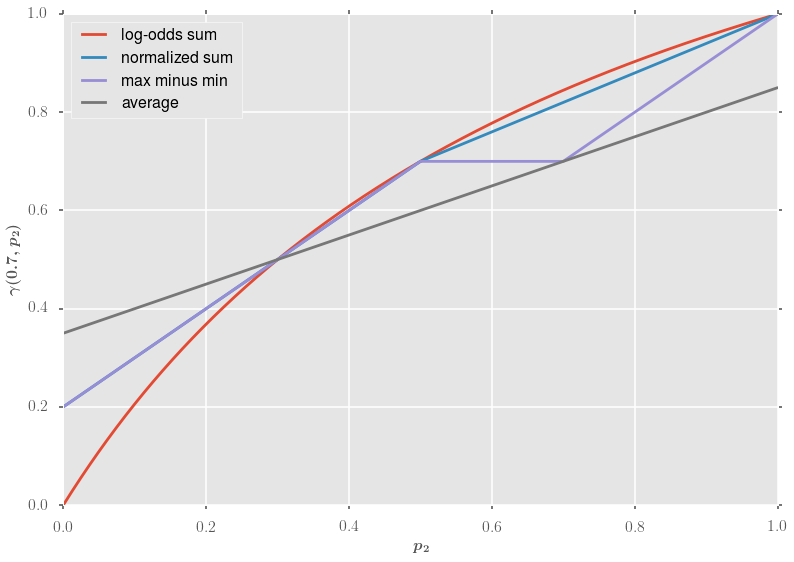}
\includegraphics[width=\textwidth]{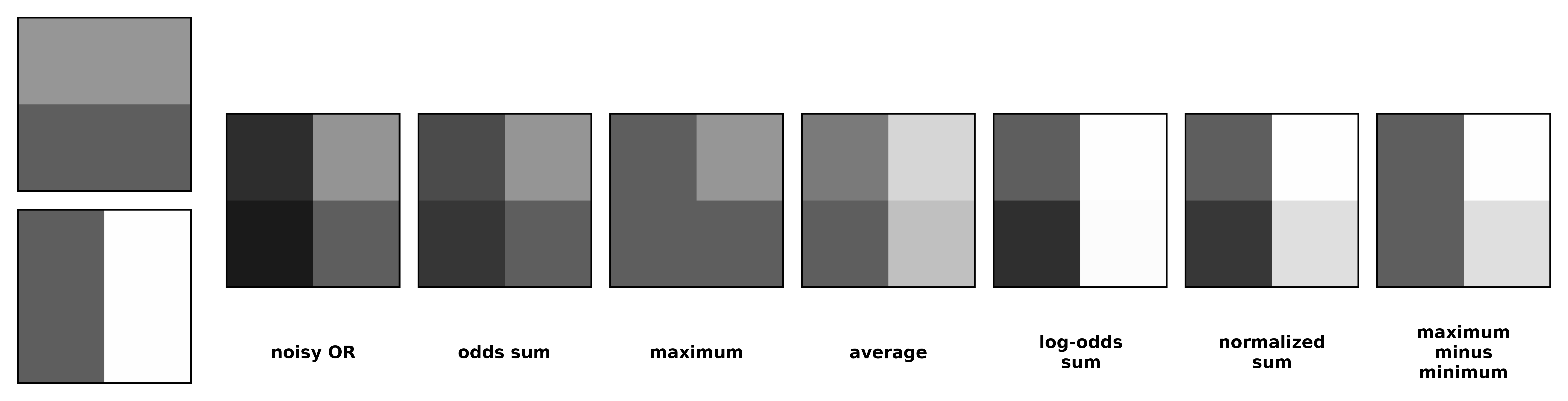}
\caption{Top: Comparison of composition rules, plotted as functions of $p_2$ for $p_1 = 0.7$. Bottom: Compositions of two experts (the probabilities in the first template are 0.5 and 0.7, the probabilities in the second template are 0.7 and 0.01) using different rules.}
\label{fig:composition_rules}
\end{figure}

The left panel in Figure \ref{fig:composition_rules} visualizes the different asymmetric composition rules for two experts. We see that the max rule is flat at 0 and hence inhibits the experts from leaving the ``no vote'' state.

\subsection{Symmetric rules}

\subsubsection{Arithmetic mean}
An intuitive composition rule for write-white-and-black models is the simple average
\[
\gamma(p_1,\ldots,p_K) = \frac{1}{K} \sum_{k=1}^K p_k.
\]
The composition can be interpreted as an equal mixture of the individual expert templates. This rule was used by \citet{amit2007pop}. Note however that with this composition rule it is impossible for an expert to abstain from voting. Consequently, the support of the experts has to be restricted manually.

\subsubsection{Sum of log-odds}
An often used composition rule is the sum of log-odds
\[
\gamma(p_1,\ldots,p_K) = \sigma\left(\sum_{k=1}^K \log \frac{p_k(x)}{1-p_k(x)}\right),
\]
where $\sigma(t) = (1+e^{-t})^{-1}$ is the logistic function (the inverse of the logit function). This type of composition is used in restricted Boltzmann machines \citep{hinton2002training} and sigmoid belief networks \citep{neal1992connectionist}. A sum followed by the logistic link function is also used in generalized linear models for binary data. This includes logistic PCA \citep{schein2003generalized}, latent trait models \citep{bartholomew2011latent}, exponential family sparse coding \citep{lee2009exponential} and binary matrix factorization \citep{meeds2006modeling}. Abstention is expressed through log-odds of 0, i.e., probabilities of 1/2. Since the first step is to sum up the individual votes (casted in terms of log-odds), opinions of experts voting in the opposite direction can be completely canceled out. Indeed, as seen in Figure \ref{fig:composition_rules}, even if $p_1 = 0.7$ the composed probability approaches 0 as $p_2$ approaches 0. As a result, there is little pressure to abstain from voting. At the same time, two very similar experts may complement each other without reducing the likelihood compared to a single strong expert.

\subsubsection{Normalized sum}
A more competitive composition rule was proposed by \citet{saund1995multiple}, which for active disagreement results in a net uncertainty. For $p_k \in \{ 0,1/2,1 \}$ the rule is
\[
\gamma(p_1,\ldots,p_K) = \frac{1}{2} \left( \frac{\sum_{k=1}^K (p_k - \frac{1}{2})}{\sum_{k=1}^K |p_k - \frac{1}{2}|} + 1 \right)
\]
and linear interpolation is used for values in between. However, due to the computational cost for linear interpolations in $K$ dimensions a more tractable approximation was actually used for the experiments in that work.

\subsubsection{Maximum minus minimum}
We propose a new symmetric composition rule, which reduces redundancy among experts and incentivizes abstention at the same time. Analogously to the max composition, we would like to rule out the possibility to increase the likelihood just by using the same expert multiple times. This can be achieved by using only the most extreme opinion. On the other hand, similarly to the normalized sum, opposing opinions should result in a limitation of the maximum achievable likelihood. These considerations naturally lead to the max-minus-min rule
\[
\gamma(p_1,\ldots,p_K) = q + \left(\smashoperator[r]{\max_{k=1,\ldots,K}} \, p_k-q\right)_+ - \left(\smashoperator[r]{\min_{k=1,\ldots,K}} \, p_k-q\right)_-.
\]
The subscript $+/-$ denotes the positive/negative part of a real number. We set $q = 1/2$, but other values could also be used. To our knowledge, this composition rule has not been used before.

\begin{figure}
\centering
\includegraphics[width=\textwidth]{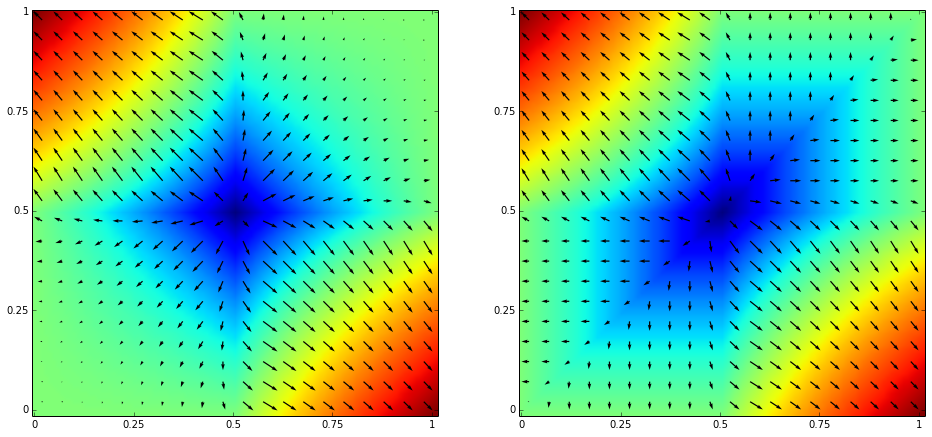}
\caption{Log-likelihood functions for a two-expert model with sum-of-log-odds composition (left) and max-minus-min composition (right).}
\label{fig:loglik_fcts}
\end{figure}

A comparison of the symmetric composition rules is shown in the right panel of Figure \ref{fig:composition_rules}. To illustrate the difference between the sum of log-odds composition and the max-minus-min composition we compute the corresponding log-likelihood functions in a simple example. Consider the ground-truth model, which creates completely white images with probability 1/4, completely black images with probability 1/4 and random images (in which each pixel is drawn independently from a Bernoulli-1/2 distribution) with probability 1/2. We attempt to learn this model by training two experts, which can be combined using the sum-of-log-odds or the max-minus-min composition rule, respectively. For simplicity we reduce the expert templates to a single parameter $\bm{\mu_k}(x) = p_k$, $k=1,2$, i.e., each pixel has the same chance of being turned on. We show the resulting (expected) log-likelihood functions in Figure \ref{fig:loglik_fcts} in the limit of a large image resolution. There are two equivalent global maxima, one at $(1,0)$ and one at $(0,1)$, each of them corresponds to one expert for black images and one expert for white images. If the initial parameters in the sum-of-log-odds model are on the same side of 1/2 (top-right or bottom-left quadrant) gradient descent will change both parameters in the same direction (i.e., it will increase both values or it will decrease both values). This partially explains why restricted Boltzmann machines \citep{hinton2002training} with randomly initialized weights have a tendency to yield multiple similar experts. The situation is very different for the max-minus-min composition. If both parameters are on the same side of 1/2 then moving along the gradient direction would only change the more extreme value while the other expert would remain close to the ``no vote'' state (and would hence be ``available'' in later stages of the learning process). Note however that we are not using gradient descent to train max-minus-min models but rather employ the EM algorithm (see Section \ref{sec:learning}). The example also suggests that it could be beneficial to use a sequential initialization procedure in which additional experts take care of structures that cannot be explained by the existing ones.

\section{Sequential inference}
\label{sec:sequential_inference}

\begin{figure}
\centering
\includegraphics[width=\textwidth]{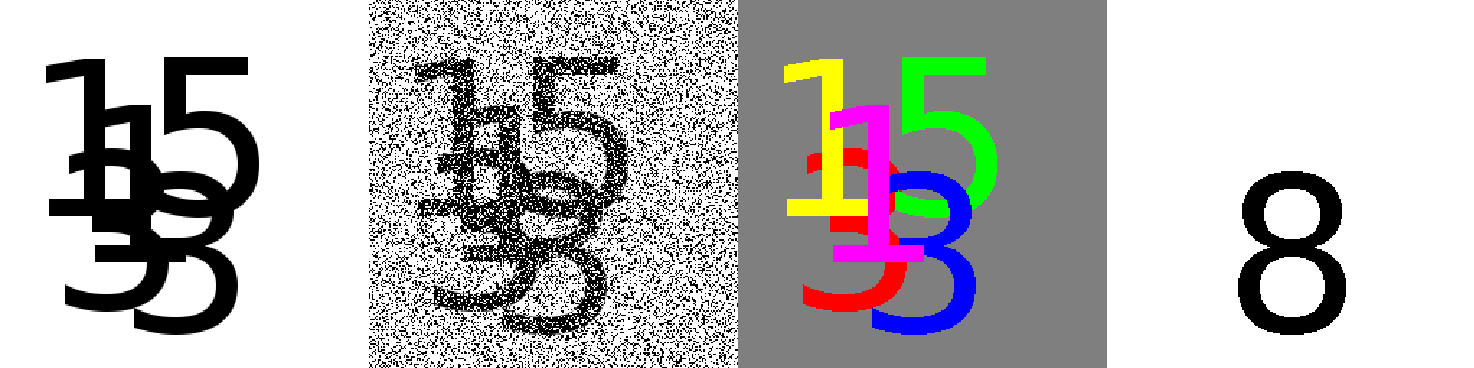}
\caption{1st panel: Scene to be analyzed. 2nd panel: Noisy version. 3rd panel: Resolved scene using robustified templates (the digits are detected in the order red, green, blue, yellow, magenta). 4th panel: First detected digit using the original templates.}
\label{fig:scene_analysis}
\end{figure}

Given experts $\bm{\mu_k}$ the inference task is to determine the posterior distribution on expert configurations given the observation $\bm{x}$, or at least finding the expert configuration that is most likely to have generated the data. For our purposes it is important that the inference procedure yields compact representations. We mention a few common approaches before we present a sequential procedure that we call \emph{likelihood matching pursuit}. In restricted Boltzmann machines \citep{hinton2002training} experts are evaluated independently. This is computationally efficient but activates all experts that match the data sufficiently well, rather than providing the sparsest possible activation that can explain the data. In other works, the indicator variables for expert presence are relaxed to real values in $[0,1]$. For example, \citet{dayan1995competition} and \citet{vincent2008extracting} use simple mean-field approximations. \citet{saund1995multiple} uses gradient descent starting from a point with all coordinates equal to 1/2. This iterative procedure becomes inefficient however if the number of possible experts is much larger than the typical number of active experts. \citet{lucke2008maximal} use a truncated search, which evaluates all expert configurations with a small number of active components.

We propose a simple sequential inference procedure for max and max-minus-min compositions. The data is first explained by the expert $\bm{\mu} = \bm{\mu_{k_1}}$ that yields the highest likelihood for the observation $\bm{x}$. Additional experts $\bm{\mu_k}$ are then evaluated by forming the composition (according to the chosen rule) of the current global template $\bm{\mu}$ with the candidate experts and computing the new likelihood. The best expert is added and the global template is updated accordingly. The procedure ends when the likelihood cannot be improved anymore. This yields a sparse activation in a single sequential pass because experts are only added if they are able to explain structures that have not been explained before. Note that this procedure is quite similar to matching pursuit \citep{mallat1993matching} for sparse coding. Instead of minimizing the squared error, we maximize the Bernoulli likelihood. This sequential fitting works well for write-white-and-black models (using the max-minus-min rule). However, since write-black models (using the max rule) encode abstention through a probability of $0$ rather than 1/2, the value of $P(\bm{x} \,|\, \bm{\mu_k})$ will heavily depend on which structures other than the one described by $\bm{\mu_k}$ are present in the observation $\bm{x}$. Consequently, it is easy to run into situations where
\[
P(\bm{x} \,|\, \bm{\mu_{k_1}}) > P(\bm{x} \,|\, \bm{\mu_{k'_1}})
\]
but
\[
P(\bm{x} \,|\, \gamma(\bm{\mu_{k_1}},\bm{\mu_{k_2}},\ldots,\bm{\mu_{k_J}})) \ll P(\bm{x} \,|\, \gamma(\bm{\mu_{k'_1}},\bm{\mu_{k_2}},\ldots,\bm{\mu_{k_J}})).
\]
Hence, the choice of the first expert may lead to a poor local maximum. The problem is that the first expert tries to explain the entire data vector. As a simple fix we propose to use truncated templates
\[
\bm{\tilde{\mu}_k}(d) = \max\left(\frac{1}{2}, \, \bm{\mu_k}(d)\right)
\]
instead. This eliminates the impact of the data in the background region of the expert. Indeed, $P(\bm{x} \,|\, \bm{\tilde{\mu}_k})$ only depends on variables $\bm{x}(d)$ for those $d$ that satisfy $\bm{\mu_k}(d) > 1/2$. That means the likelihood of the truncated template only depends on the data in the support of expert $k$. A similar idea to robustify templates was used by \citet{williams2004greedy}. They mix the original templates with a uniform distribution. In the case of binary data this amounts to $\alpha \bm{\mu}(d) + (1-\alpha)/2$ for some $\alpha \in (0,1)$, i.e., a convex combination between $\bm{\mu}$ and 1/2 is formed. For a write-black model this is a less effective transformation than our truncation. We illustrate the effectiveness of the proposed modification through a simple scene analysis example. Five digits are placed at random locations in the image according to a write-black model and the task is to recover the identity and locations of the digits in a noisy version of this scene. Using the robustified templates the scene can be resolved perfectly, as shown in Figure \ref{fig:scene_analysis}. However, with the original templates the first digit, which is placed down, corresponds to the largest structure in the image that can be explained by a single template through a moderately well fit. Note that the suggested fix is not the only possible solution. If we allow for an iterative procedure then the scene can also be resolved with the original templates. The modification is simply meant as a robustification of the greedy inference procedure.

\section{EM learning}
\label{sec:learning}
We now describe an approximate EM procedure \citep{dempster1977maximum} for learning max-minus-min models. When working with image data, we explicitly model transformations like shifts and rotations. Each template $\bm{\mu_k}$ then provides multiple transformed versions $\bm{\mu_{kt}} = \Phi_t(\bm{\mu_k})$ where $t$ denotes the transformation. This allows us to share parameters among all transformed versions. Since our templates will describe rather large structures, we assume that only one of the transformed versions is present in each image. In the (hard) E-step we use the likelihood matching pursuit procedure from the previous section, which yields for each observation $\bm{x}$ a representation $(k_1,t_1),\ldots,(k_J,t_J)$. We define
\[
k^\star(d) = \argmax_{j=1,\ldots,J} \bm{\mu_{k_jt_j}}(d), \qquad \ell^\star(d) = \argmin_{j=1,\ldots,J} \bm{\mu_{k_jt_j}}(d)
\]
provided that the maximum is larger than $q$ and the minimum is smaller than $q$, respectively (otherwise we leave $k^\star(d)$ or $\ell^\star(d)$ undefined). In words, $k^\star(d)$ and $l^\star(d)$ are the active experts with the most extreme opinions for variable $\bm{x}(d)$. In the M-step we then simply compute
\begin{equation}
\bm{\mu_k}(d) = \frac{\sum_{n=1}^N \mathbbm{1}\{k{=}k_n^\star(d) \text{ or } k {=}\ell_n^\star(d)\} \, \Phi_{t_{nk}}^{-1}(\bm{x_n})(d) + \epsilon}{\sum_{n=1}^N \mathbbm{1}\{k{=}k_n^\star(d) \text{ or } k{=}\ell_n^\star(d)\} + 2\epsilon}
\label{eq:M_step}
\end{equation}
where $k_n^\star$, $l_n^\star$ are the maximizers respectively minimizers and $t_{nk}$ is the transformation corresponding to $k$-th expert in the representation of the $n$-th training example $\bm{x_n}$. The pseudocount $\epsilon > 0$ is added for regularization purposes and can be interpreted as a Beta$(\epsilon,\epsilon)$ prior on the expert probabilities. In all our experiments we use $\epsilon=1$, which corresponds to a uniform prior. The update formula (\ref{eq:M_step}) is not the exact maximizer for the expert templates but a very good heuristic. The reason why we can use such a simple analytic expression is that with extremal composition rules the responsibility for individual variables $\bm{x}(d)$ is not split among many experts. With other composition rules gradient descent methods are often required in order to perform the M-step. Also note that for $q=0$ the max-minus-min model reduces to the max model. In that special case the update formula (\ref{eq:M_step}) only involves $k^\star$ but not $\ell^\star$.

\subsection{Online learning and sequential initialization}
\label{subsec:online_learning}
It is straightforward to provide an online version of the learning procedure. The M-step updates simply are
\begin{gather*}
\bm{\mu_k}(d) \leftarrow \frac{N_k(d) \bm{\mu_k}(d) + \mathbbm{1}\{k{=}k_n^\star(d) \text{ or } k{=}\ell_n^\star(d)\} \, \Phi_{t_{nk}}^{-1}(\bm{x_n})(d)}{N_k(d) + \mathbbm{1}\{k{=}k_n^\star(d) \text{ or } k{=}\ell_n^\star(d)\}},\\
\\
N_k(d) \leftarrow N_k(d) + \mathbbm{1}\{k{=}k_n^\star(d) \text{ or } k{=}\ell_n^\star(d)\}.
\end{gather*}
The variable $N_k(d)$ counts how many training examples have been used to compute the current estimate for the $d$-th dimension of expert $k$. The online version is attractive because it allows us to use a sequential initialization scheme. Since the learning problem is non-convex, it is crucial to have a good initialization. In accordance with our attempt to learn a parsimonious representation we start off with a single global template derived from the first training example and add more experts later on. The idea is to use ``oversimplified'' models in the sense that they try to explain new examples through the experts learned so far. The models are then ``corrected'' by appending additional templates to the collection of experts. Define $\bm{\tilde{\mu}_{K+1}}(d) = (\bm{x_n}(d)+\epsilon) / (1+2\epsilon)$. When using the max-minus-min composition the additional template can be initialized as
\[
\bm{\mu_{K+1}}(d) = 
\begin{cases}
\frac{1}{2} & \textrm{if } \mathbb{P}(\bm{x_n}(d) \,|\, \bm{\mu}(d)) \geq \mathbb{P}(\bm{x_n}(d) \,|\, \bm{\tilde{\mu}_{K+1}}(d))\\
\bm{\tilde{\mu}_{K+1}}(d) & \textrm{otherwise}
\end{cases}
\]
where $\bm{\mu}$ is the composed template using the existing experts. This means that the new expert abstains from voting for dimensions that are already well explained by the other experts. For max compositions we suggest to use
\[
\bm{\mu_{K+1}}(d) = 
\begin{cases}
\frac{1}{2} & \textrm{if } \mathbb{P}(\bm{x_n}(d) \,|\, \bm{\mu}(d)) \geq \frac{1}{2}\\
\bm{\tilde{\mu}_{K+1}}(d) & \textrm{otherwise}
\end{cases}.
\]
This makes sure that in the background region $\bm{\mu_{K+1}}$ is close to 0 (rather than equal to 1/2). Our successive refinement is in stark contrast to the typical bottom-up grouping of local structures in part-based compositional models. For example, \citet{fidler2007towards} and \citet{zhu2008unsupervised} start with elementary edge features and combine simple structures into more complex ones through a hierarchical clustering procedure.

\subsection{Geometric component}
For image data, the spatial arrangement of the experts can be modeled through a joint Gaussian distribution for shifts and rotations. This only requires us to compute the mean and covariance of the training configurations provided by the inference procedure. As emphasized by \citet{bruna2013invariant}, a very desirable property of a representation is that intraclass deformations are linearized. As the experiment in Section \ref{subsec:handwritten_letters} confirms, our representation transforms the complex deformation orbit (in the original space) into a linear space in which a Gaussian distribution satisfactorily describes the deformations.

\section{Experiments}
\label{sec:experiments}

\subsection{A synthetic write-white-and-black model}

\begin{figure}
\centering
\includegraphics[width=\textwidth]{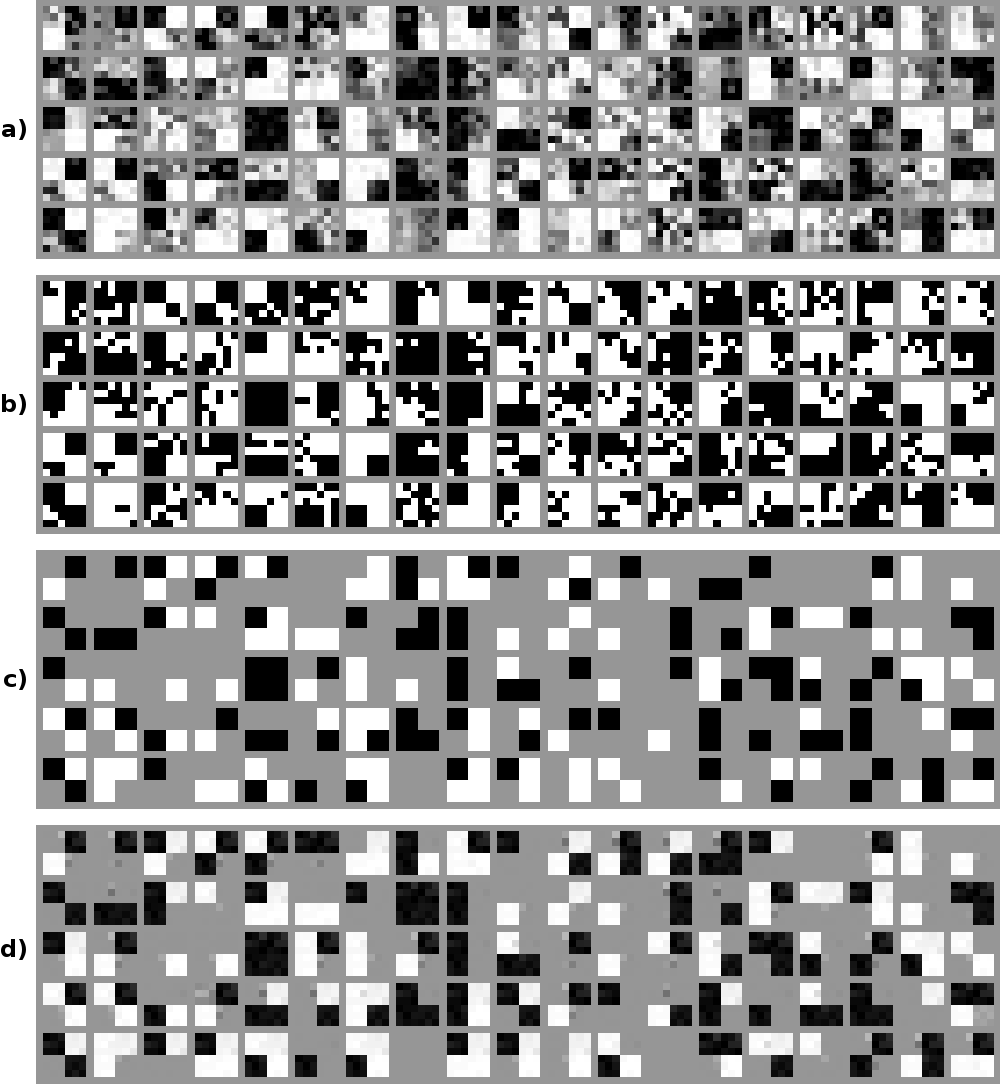}
\caption{a) Sparse coding reconstruction. b) 100 samples from the synthetic model. c) Ground-truth templates for the samples above. d) Reconstruction of the max-minus-min model.}
\label{fig:synthetic_reconstruction}
\end{figure}

\begin{figure}
\centering
\includegraphics[width=.475\textwidth]{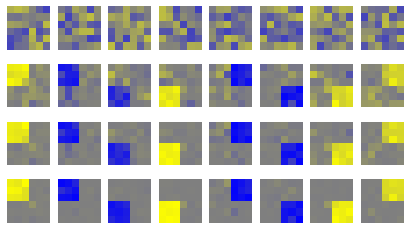}
\hspace{.03\textwidth}
\includegraphics[width=.475\textwidth]{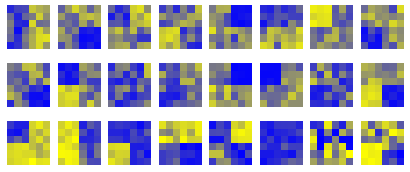}
\caption{Left: Initialization (1st row) and learned experts after 1, 2 and 5 EM iterations (2nd - 4th row) for the max-minus-min model trained on 100 examples from the synthetic model. Blue, gray and yellow color corresponds to probabilities of 0, 1/2 and 1, respectively. Right: Experts obtained from a denoising autoencoder (top), a restricted Boltzmann machine (center) and dictionary learning for sparse coding (bottom) using 100 samples.}
\label{fig:synthetic_templates}
\end{figure}

\begin{figure}
\centering
\includegraphics[width=.67\textwidth]{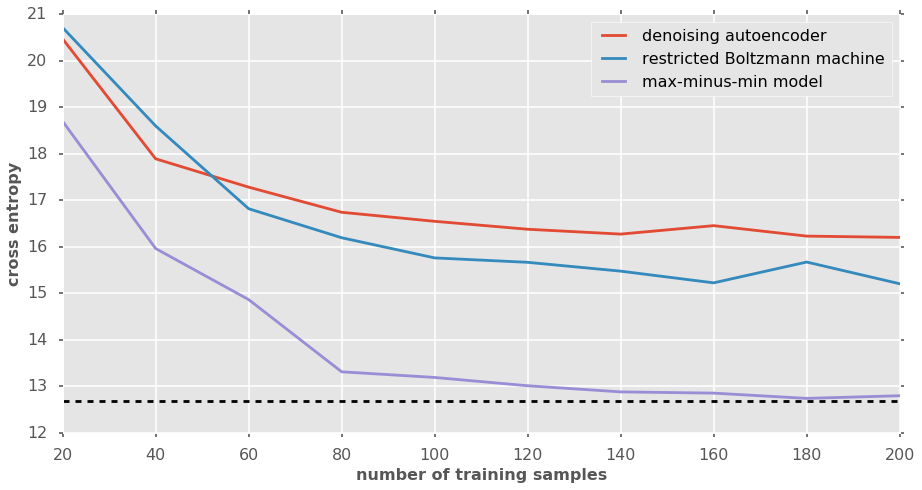}
\caption{Cross-entropy reconstruction error for different models and training sizes (lower is better). The dashed black line is the cross-entropy of the ground-truth model.}
\label{fig:synthetic_entropy}
\end{figure}

We compare the max-minus-min model to denoising autoencoders \citep{vincent2008extracting} and restricted Boltzmann machines \citep{hinton2002training} on a synthetic model for binary images of size 6 $\times$ 6 pixels. The image grid is divided into a regular grid of four quadrants. Each quadrant is independently activated with probability 1/2. An activated quadrant is either entirely black or entirely white, each with probability 1/2. For non-activated quadrants the pixels are drawn independently from a Bernoulli-1/2 distribution (samples from this model can be found in Figure \ref{fig:synthetic_reconstruction}). The task is to recover the 8 underlying experts corresponding to the four quadrants with two polarities. The learning algorithm for the max-minus-min model typically converged after a few iterations. In the left panel of Figure \ref{fig:synthetic_templates} we visualize the obtained templates after 1, 2 and 5 iterations when using 100 training examples. As we see, our model is able to almost perfectly recover the original experts. The denoising autoencoder and the restricted Boltzmann machine were run for 100 epochs and the learning rates were tuned. For the denoising autoencoder we also tuned the corruption level. Figure \ref{fig:synthetic_entropy} shows a comparison with the max-minus-min model in terms of the cross-entropy (i.e., the negative log-likelihood of the composed templates) on a test set of 1,000 samples. The reconstruction error for the max-minus-min model is much lower. This is because with the sum-of-log-odds composition multiple ground-truth experts are mixed up. The right panel of Figure \ref{fig:synthetic_templates} shows that the templates are indeed combinations of multiple ground-truth experts, i.e., the factors of variation have not been disentangled successfully. In order to make a fair comparison we have left out the sequential initialization procedure (as well as the transformation modeling) and initialized all models with completely random templates. That means the performance gain is in fact due to the more competitive expert interaction. As a further comparison we learned a sparse coding dictionary \citep{mairal2009online}. Note that the data is then treated as real-valued and the task is to find basis vectors whose linear combinations allow for good reconstructions in a squared error sense. The basis vectors learned from 100 samples are visualized in the right panel of Figure \ref{fig:synthetic_templates}. Again, the ground-truth experts are not recovered. However, the basis vectors look more structured than the experts learned by denoising autoencoders or restricted Boltzmann machines. In terms of $L_2$ reconstruction error this dictionary is actually even better than the ground-truth generative model. Figure \ref{fig:synthetic_reconstruction} visualizes the reconstructions obtained via sparse coding and via the learned max-minus-min model. The sparse coding reconstruction is visually closer to the data but much more noisy than the reconstruction of our model. Indeed, the max-minus-min reconstruction is almost identical to the ground-truth templates. The reason for this different behavior is that the squared error is more forgiving for small deviations compared to cross-entropy and penalizes harder than cross-entropy if the deviation is large (around 1/2).

\subsection{Handwritten letters}\label{subsec:handwritten_letters}

\begin{figure}
\centering
\includegraphics[height=.125\textheight]{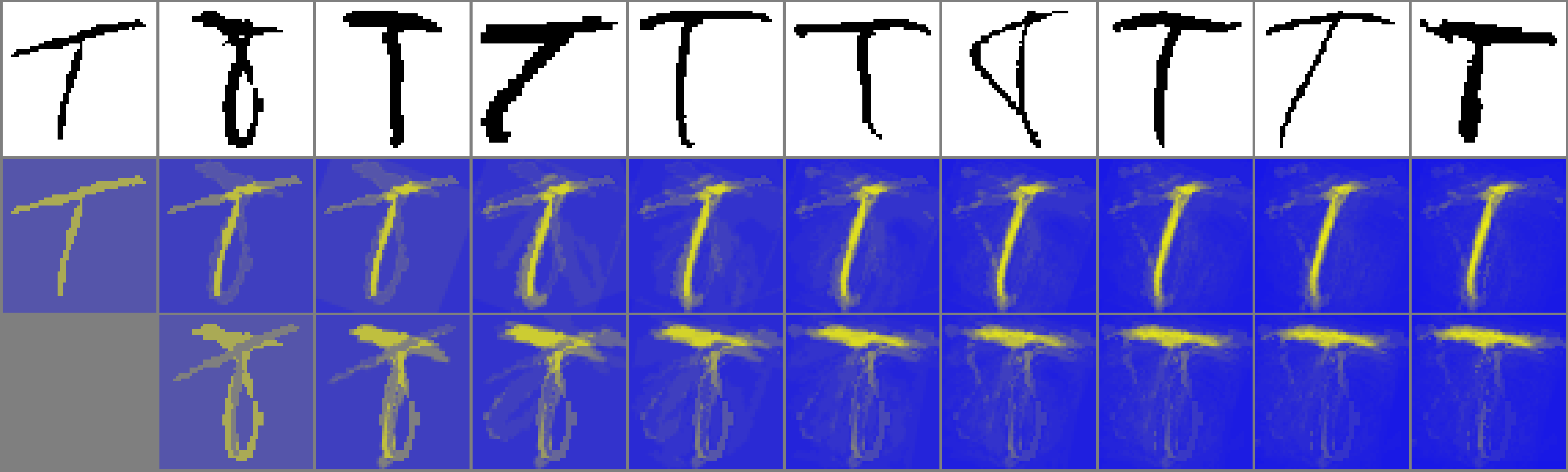}
\hspace{2em}
\includegraphics[height=.125\textheight]{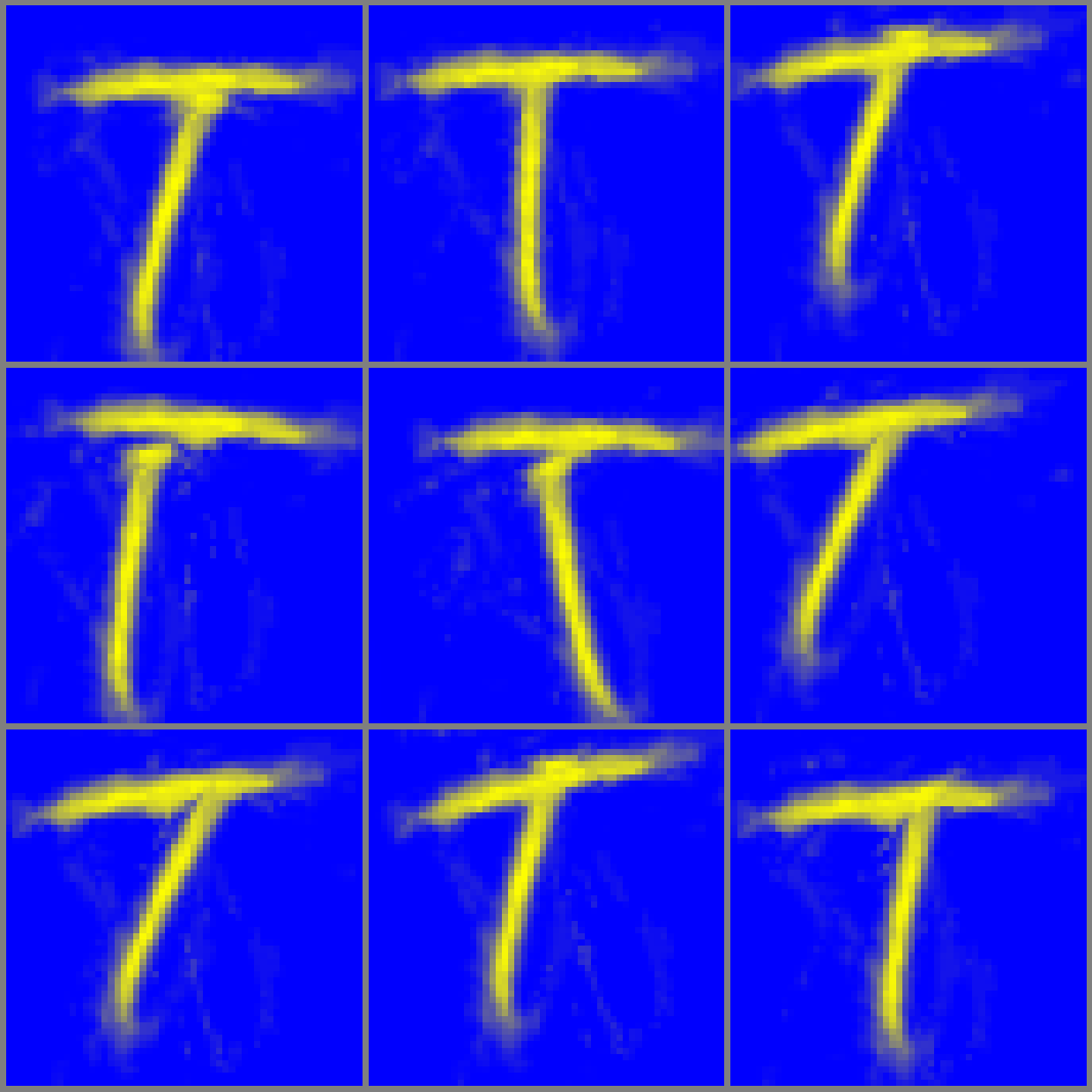}
\caption{Online learning for the letter T. Left panel: The 10 samples used for training (1st row) and the two expert templates (2nd \& 3rd row) at step $i=1,\ldots,10$. Right panel: 9 sampled expert configurations based on a multivariate Gaussian distribution of the spatial expert arrangement using the final templates.}
\label{fig:learn_and_sample_t}
\end{figure}

\begin{figure}
\centering
\includegraphics[width=.5\textwidth]{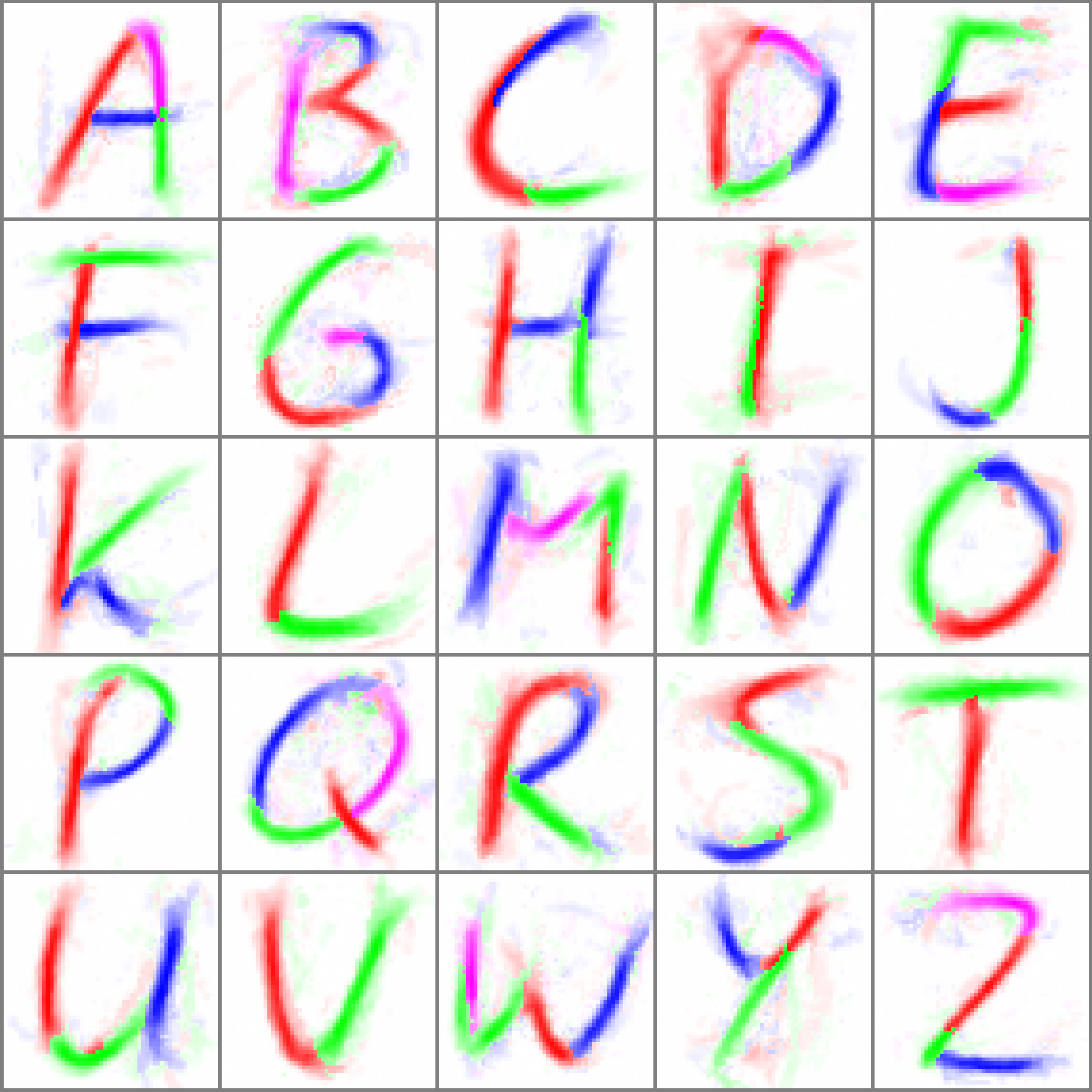}
\caption{Experts learned from 20 examples per letter class. Each expert is plotted at its mean location with mean orientation. For each pixel the color (red, green, blue, magenta) indicates the maximum expert and the intensity visualizes the template value (white corresponding to 0, color corresponding to 1).}
\label{fig:letters}
\end{figure}

We now train experts on the letter classes from the TiCC handwritten characters dataset \citep{vandermaaten2009new} while modeling shifts and rotations. The images are of size $56 \times 56$ pixels and each sample provides a label for the writer of the corresponding character. Since ``on'' pixels in this dataset correspond to black ink it is natural to use the max composition rule. The purpose of this experiment is to illustrate the effectiveness of our sequential initialization procedure from Section \ref{subsec:online_learning}. The left panel in Figure \ref{fig:learn_and_sample_t} visualizes the online learning process for the letter T, using the first sample of the first 10 writers. The first expert is initialized by the first training example. The second expert is initialized by the characteristics of the second example that cannot be explained through the first expert. Every additional image then updates both experts. After 10 examples the learning process has converged to a vertical and a horizontal bar. Samples (Figure \ref{fig:learn_and_sample_t}, right panel) from the learned spatial distribution look realistic and cover the principal deformations of the class. We also learned up to four experts for the other\footnote{The letter X is missing from the dataset.} letters, using the first sample of the first 20 writers. The results are shown in Figure \ref{fig:letters}. Most experts correspond to natural elements of the character class. Note that in contrast to \citet{lake2013one} no motion information was necessary to learn these character primitives.

\section{Conclusion and future work}
We considered various composition rules and discussed their adequacy for learning compact representations. Inference and learning procedures for models with extremal composition rules were then proposed and their performance was tested in experiments. An alternative to competitive interaction rules would be to use a prior on the part parameters (e.g., an $L_1$-penalty on the log-odds). However, this would create a bias, which affects all experts. Our approach on the other hand allows us to use maximum-likelihood estimation.

The focus of this paper was on binary data. A natural next step is to study compositions of experts for real-valued data. This includes considering composition rules for variances (in addition to means) and achieving continuity when transitioning from one expert to another.

\bibliography{iclr2015}
\bibliographystyle{iclr2015}

\end{document}